\documentclass[letterpaper, 10 pt, conference]{ieeeconf}  % Comment this line out if you need a4paper

\IEEEoverridecommandlockouts                              % This command is only needed if 
\usepackage{amssymb,amsmath}
\usepackage{booktabs}
\usepackage{caption}
\usepackage{subcaption}
\usepackage{graphicx}
\usepackage{graphics}

\usepackage{multicol,multirow}
\usepackage{verbatimbox}

\usepackage{hyperref}
\hypersetup{bookmarksopen,bookmarksnumbered,
pdfpagemode=UseOutlines,
colorlinks=true,
linkcolor=blue,
anchorcolor=blue,
citecolor=blue,
filecolor=blue,
menucolor=blue,
urlcolor=blue
}

\title{\LARGE \bf
TBD Pedestrian Data Collection: Towards Rich, Portable, and Large-Scale Natural Pedestrian Data 
}

\author{Allan Wang$^{1}$, Daisuke Sato$^{1}$, Yasser Corzo$^{1}$, Sonya Simkin$^{1}$, Abhijat Biswas$^{1}$, and Aaron Steinfeld$^{1}$% <-this % stops a space
\thanks{$^{1}$The authors are with the Robotics Institute, Carnegie Mellon University,
        5000 Forbes Avenue, Pittsburgh, PA, USA 
        {\tt\small \{allanwan, daisukes, ycorzo, ssimkin, abhijatb\}@andrew.cmu.edu, 
        steinfeld@cmu.edu}}%
}

\begin{document}

\maketitle
\thispagestyle{empty}
\pagestyle{empty}

%%%%%%%%%%%%%%%%%%%%%%%%%%%%%%%%%%%%%%%%%%%%%%%%%%%%%%%%%%%%%%%%%%%%%%%%%%%%%%%%
\begin{abstract}

Social navigation and pedestrian behavior research has shifted towards machine learning-based methods and converged on the topic of modeling inter-pedestrian interactions and pedestrian-robot interactions. For this, large-scale datasets that contain rich information are needed. We describe a portable data collection system, coupled with a semi-autonomous labeling pipeline. As part of the pipeline, we designed a label correction web application that facilitates human verification of automated pedestrian tracking outcomes. Our system enables large-scale data collection in diverse environments and fast trajectory label production. Compared with existing pedestrian data collection methods, our system contains three components: a combination of top-down and ego-centric views, natural human behavior in the presence of a socially appropriate ``robot", and human-verified labels grounded in the metric space. To the best of our knowledge, no prior data collection system has a combination of all three components. We further introduce our ever-expanding dataset from the ongoing data collection effort -- the \textit{TBD Pedestrian Dataset} and show that our collected data is larger in scale, contains richer information when compared to prior datasets with human-verified labels, and supports new research opportunities.

\end{abstract}

%%%%%%%%%%%%%%%%%%%%%%%%%%%%%%%%%%%%%%%%%%%%%%%%%%%%%%%%%%%%%%%%%%%%%%%%%%%%%%%%
\section{Introduction}

Pedestrian datasets are essential tools for modeling socially appropriate robot behaviors, recognizing and predicting human actions, and studying pedestrian behavior. Researchers may use these data to predict future pedestrian motions, including forecasting their trajectories \cite{Alahi1, Gupta1, ivanovic-trajectron}, and/or navigation goals \cite{kitani-2012, liang2020garden}. In social navigation, datasets can also be used to model \cite{okal-IRL, kretzschmar_ijrr16} or evaluate robot navigation behavior \cite{biswas2021socnavbench}. For this, an in-the-wild pedestrian dataset that is large-scale and supports ground-truth metric labels is desired.

However, existing public pedestrian datasets are either unlabelled \cite{karnan2022scand, diego2022crowdbot}, only rely on labels produced by an automated pipeline \cite{atc, edinburgh}, only contain pixel level information \cite{stanforddrone, virat} or are small in scale \cite{ETH, UCY, jrdb}. We propose a system that can collect large quantities of quality data efficiently. The data collected using our system feature a novel full combination of three critical elements: a combination of top-down and ego-centric views, natural human motion, and human-verified labels grounded in the metric space. This allows the data collected using our system to contain rich information.

\begin{figure}[]
      \centering
      \includegraphics[scale=0.5]{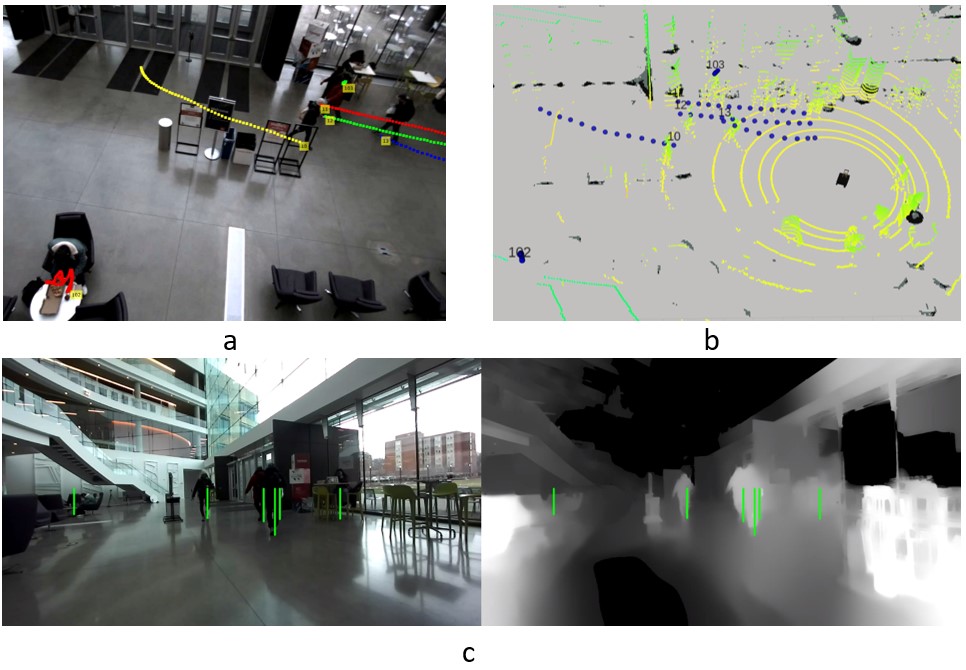}
      \caption{This set of images represent the same moment recorded from multiple sensors: a) Top-down view image taken by a static camera with grounded pedestrian trajectory labels shown. b) Ego-centric point cloud from a 3D LiDAR with the projected trajectories from (a). c) Ego-centric RBG and depth images from the mounted stereo camera. Green vertical bars represent the projected labels. Note that two pedestrians at the back are partially and completely occluded from the stereo camera.}
      \label{fig:intro}
      \vspace{-1em}
  \end{figure}

Large datasets with quality labels and rich information can assist in addressing human behavioral research questions that require the modeling of interaction. For example, a key problem researchers have tried to address is the \textit{freezing robot problem} \cite{Trautman1}. Researchers have attributed this problem to the robot's inability to model interactions \cite{sun2021move}. Works such as \cite{nishimura2020risk} have used datasets to show that modeling the anticipation of human reactions to the robot's actions enables the robot to deliver a better performance. However, interactions are diverse and uncommon in human crowds, they contain many types \cite{mavrogiannis_etal2021-core-challenges} (e.g. passing, following, grouping) and can further be diversified by the environment (e.g. an open plaza or a narrow corridor), so pedestrian datasets need to be large-scale in order to capture enough interaction data. 

Autonomous vehicle datasets \cite{nuscenes, Cityscapes} have inspired a plethora of research. However, a dataset of similar caliber and label quality in pedestrian-dominant environments has yet to arrive. As a step toward this goal, we have constructed a data collection system that can achieve these two requirements: large data quantity and diversity, and human-verified positional labels. First, we ensure that our equipment is portable and easy to set up. This allows collecting data in a variety of locations with limited lead time. Second, we address the challenge of labeling large quantities of data using a semi-autonomous labeling pipeline. We employ a state-of-the-art deep learning-based \cite{zhang2021bytetrack} tracking module combined with a human inspection and tracking error-fixing web application to semi-automatically produce high-quality ground truth pedestrian trajectories in metric space. We make the web application open-source\footnote{\href{https://github.com/CMU-TBD/tbd_label_correction_UI}{https://github.com/CMU-TBD/tbd\_label\_correction\_UI}} so that other researchers can use this tool or contribute to this effort.

While we hope our contributions support robot system improvements in the community and we aim to accommodate a wide variety of pedestrian behavior research, our dataset primarily supports human environment navigation research that requires ground truth pedestrian positional information, such as social navigation, pedestrian trajectory prediction, and ego-centric perception. Specifically, we include three important characteristics. (1) Top-down view and ego-centric views: This ensures that the robot has access to ground-truth data even with occlusions. (2) Natural human motion: The manual pushing of the inconspicuous suitcase robot mitigates the curiosity effects of nearby pedestrians. And (3) Ground truth labeling in metric space: This allows our dataset to be useful for research where positional pedestrian data are needed. To the best of our knowledge, publicly available datasets only have at most two of these characteristics. 

We demonstrate our system through a dataset collected in a large indoor space: the TBD Pedestrian Dataset\footnote{\href{https://tbd.ri.cmu.edu/tbd-social-navigation-datasets}{https://tbd.ri.cmu.edu/tbd-social-navigation-datasets}}. Our dataset contains scenes with a variety of crowd densities and pedestrian interactions. We show through our analysis that our dataset (Batch 1: 133 minutes - 1416 trajectories; Batch 2: 626 minutes - 10300 trajectories) is larger in scale and contains unique characteristics compared to prior similar datasets. This is an ongoing effort and we plan to collect additional data in more diverse locations.
\section{Related Work}

\label{sec:related-dsetuse}
With the explosion of data-hungry machine learning methods in robotics, demand for pedestrian datasets has been on the rise in recent years. One popular category of research in this domain is human trajectory prediction (e.g., \cite{Alahi1, Gupta1, ivanovic-trajectron, kitani-2012, liang2020garden, Sophie, Social-STGCNN, wang-split-merge}). Much of this research utilizes selected mechanisms to model pedestrian interactions in hopes for better prediction performance (e.g., pooling layers in the deep learning frameworks \cite{Alahi1, Gupta1} or graph-based representations \cite{Social-STGCNN}). Rudenko et al. \cite{rudenko2019-predSurvey} provides a good summary into this topic. While the state-of-the-art performance keeps improving with the constant appearance of newer models, it is often unclear how well these models can generalize in diverse environments. As shown in \cite{rudenko2019-predSurvey}, many of these models only conduct their evaluation on the relatively small-scale ETH \cite{ETH} and UCY \cite{UCY} datasets.

Another popular demand for pedestrian datasets comes from social navigation research. Compared to human motion prediction research, social navigation research focuses more on planning. For example, social navigation research uses learning-based methods to identify socially appropriate motion for better robot behavior, such as deep reinforcement learning \cite{Everett18_IROS, chen2019crowd, Chen-gaze-learn} or inverse reinforcement learning \cite{okal-IRL, Tai-IRL}. Due to the lack of sufficiently large datasets, these models often train in simulators that lack realistic pedestrian behavior. Apart from training, datasets are also increasing in popularity in social navigation evaluation due to their realistic pedestrian behavior \cite{gao2021evaluation}. Social navigation methods are often evaluated in environments using pedestrian data trajectory playback (e.g., \cite{sun2021move, trautmanijrr, cao2019dynamic, wang2022group}). However, similar to human motion prediction research, these evaluations are typically only conducted on the ETH \cite{ETH} and UCY \cite{UCY} datasets, as shown by \cite{gao2021evaluation}. These two datasets only use overhead views, and therefore lack the ego-centric view used by most robots. 

Large-scale and high-quality datasets exist for other navigation-related applications and research. Autonomous vehicle datasets such as nuScenes \cite{nuscenes}, Cityscapes \cite{Cityscapes} and ArgoVerse \cite{Argoverse} also contain pedestrian-related data. However, pedestrians often have limited appearances on sidewalks or at crosswalks. There is also no data on how pedestrians navigate indoors. Another group of similar datasets mainly supports computer vision-related research, such as MOT \cite{MOTChallenge20} for pedestrian tracking, and Stanford Drone Dataset (SDD) \cite{stanforddrone} and VIRAT \cite{virat} for pedestrian motion/goal prediction on the image level. Detailed comparisons of the characteristics between the TBD Pedestrian Dataset and similar existing datasets can be found in section \ref{sec:eval-compare}.

Simulators can fill in the role of datasets for both training and evaluation. Simulators such as SocNavBench \cite{biswas2021socnavbench}, CrowdNav \cite{chen2019crowd}, PedSIM \cite{gloor2016pedsim}, and SEAN \cite{tsoi2020sean} are in use by the research community. However, sim-to-real transfer is an unsolved problem in robotics. Apart from lack of fidelity in visuals and physics, pedestrian simulators in particular entail the additional paradox of pedestrian behavior realism \cite{mavrogiannis2021core}: If pedestrian models are realistic enough for use in simulators, why don't we apply the same model to social navigation?

\section{System Description}\label{sec:system}

\subsection{Hardware Setup}\label{sec:hardware}

\begin{figure}[]
    \centering
    \includegraphics[scale=0.32]{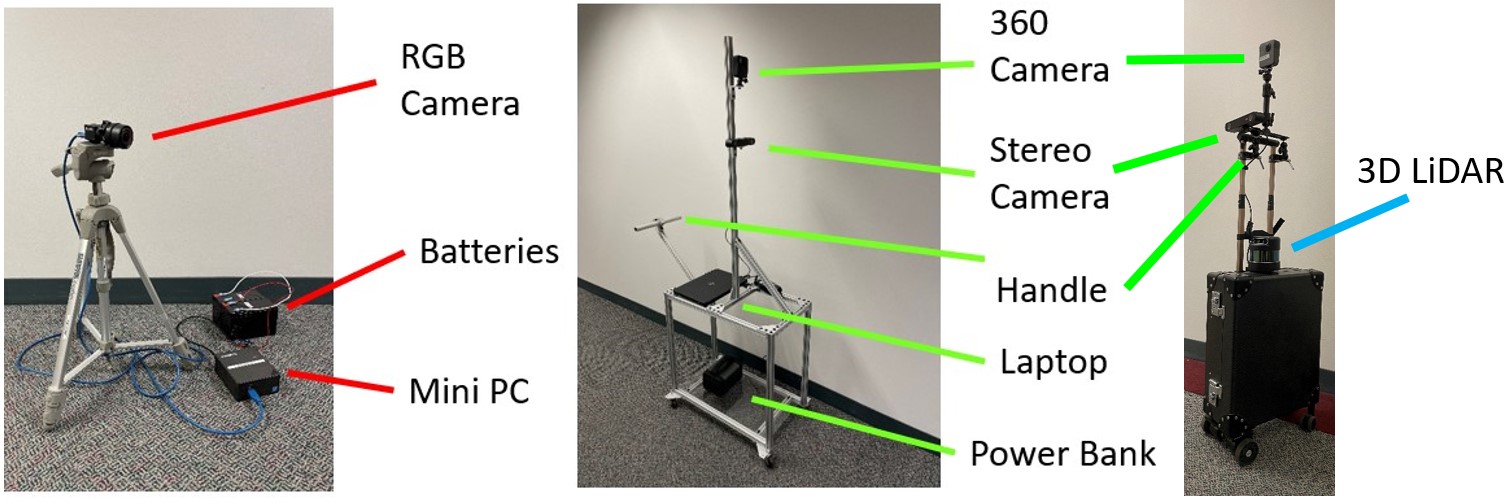}
    \caption{Sensor setup used to collect the TBD Pedestrian Dataset. (left) One of the nodes used to capture top-down RGB views. (middle) The cart used to capture ego-centric sensor views during data collection for Set 1. (right) The suitcase robot used to capture ego-centric sensor views during data collection for Set 2.}
    \label{fig:camera}
    \vspace{-1.5em}
\end{figure}

\begin{figure}[]
    \centering
    \includegraphics[scale=0.5]{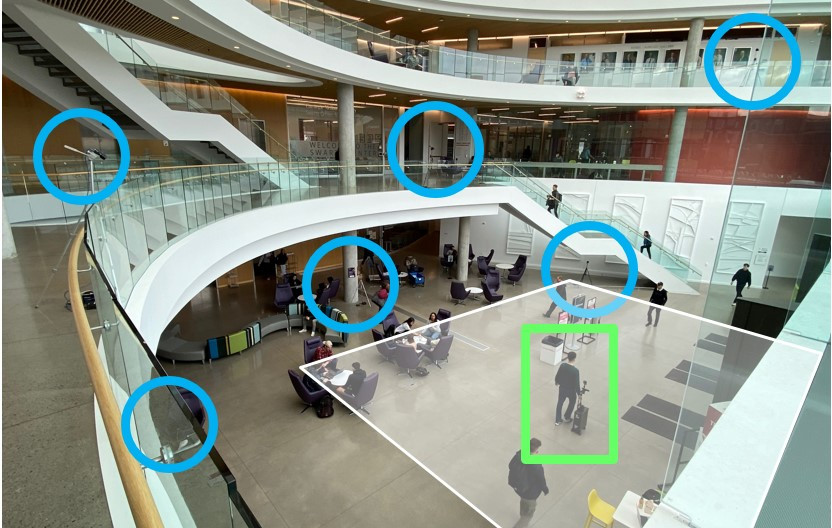}
    \caption{Hardware setup for the TBD Pedestrian Dataset. Blue circles indicate positions of RGB cameras. Green box shows our suitcase robot pushed through the scene. The white area is where trajectory labels are collected.}
    \label{fig:hdc_setup}
    \vspace{-1em}
\end{figure}

Our system supports multiple static FLIR Blackfly RGB cameras for labeling and metric space calculations (Figure~\ref{fig:camera}). The TBD Pedestrian Dataset contains two different configuration sets for the same physical space. Both sets include three cameras surrounding the scene on the upper floors overlooking the ground level at roughly 90 degrees apart from each other (Figure~\ref{fig:hdc_setup}). The RGB cameras are connected to portable computers powered by lead-acid batteries. The multiple cameras complement each other's coverage, because even from overhead views, partial occlusions occur. We used the three overhead view cameras to label trajectories. We also positioned three more units on the ground floor, but did not use them for pedestrian labeling. The cameras' intrinsics are precalibrated and their extrinsics are obtained via physical landmarks that we placed in the scene.

In addition to static cameras, we pushed a cart (Set 1) or robotic suitcase \cite{cabot2023kuribayashi} (Set 2) through the scene. The cart (Figure~\ref{fig:camera}) was equipped with a ZED stereo camera to collect ego-centric RGB-D views of the scene. A GoPro Fusion 360 camera for capturing high definition 360 videos of nearby pedestrians was mounted above the ZED. Data from the 360 camera is useful for capturing pedestrian pose data and facial expressions for future work. The ZED camera was powered by a laptop with a power bank. Our entire data collection hardware system is portable and does not require power outlets, thereby allowing data collection outdoors or in areas where wall power is inaccessible.

The robotic suitcase (Set 2) is a converted carry-on rolling suitcase. It is equipped with an IMU and a Velodyne VLP-16 3D LiDAR sensor. In addition, the same ZED camera and 360 camera are mounted on the suitcase handle. The robot's computer, batteries, and all its internal components are hidden inside the suitcase, so pushing the robot resembles pushing a suitcase. We selected this robot because of its inconspicuous design to reduce curious, unnatural reactions from nearby pedestrians, as curious pedestrians may intentionally block robots or display other unnatural movements \cite{brvsvcic2015escaping}. While it is true that real-world pedestrians will react to mobile robots curiously in the short term and some may argue in favor of a more robotic appearance, we envision that such curiosity effects will die down in the long term. 

During certain data collection sessions, we pushed the cart or the suitcase robot from one end of the scene to another end, while avoiding pedestrians and obstacles along the way in a natural motion similar to a human pushing a delivery cart or walking with a suitcase. This collects ego-centric views from a mobile robot traversing through the human environment. However, unlike other datasets such as \cite{karnan2022scand}, \cite{diego2022crowdbot}, \cite{jrdb}, and \cite{lcas} that use a tele-operated robot, or \cite{thor} that uses a scripted policy to act autonomously, we chose to have all motion performed by the human walking with the system. This provides better trajectory control, increased safety, and further reduced the novelty effect from surrounding pedestrians.

Both sets of data collection occurred on the ground level in a large indoor atrium area (Figure~\ref{fig:hdc_setup}). Half of the atrium area has fixed entry/exit points that lead to corridors, elevators, stairs, and doors to the outside. The other half of the atrium is adjacent to another large open area and is unstructured with no fixed entry/exit points. We collected data around lunch and dinner times to ensure higher crowd densities. Additional data collection are planned at more diverse locations.

\subsection{Post-processing Pipelines}

To ensure time synchronization across the captured videos, we employed Precision Time Protocol over a wireless network to synchronize each of the computers powering the cameras. In addition, we used an LED light and check for the light signal during the post-processing stage to synchronize the frames of all captured data for each recording session. 

The next step is to localize the cart/suitcase in the scene. For Set 1, this was achieved by identifying the cart/suitcase on the overhead camera videos and then applying the camera matrices to obtain the metric coordinates. For Set 2, we first made a map inside the building and then computed its location in the post processing phase by utilizing the robotic suitcase software~\footnote{\href{https://github.com/cmu-cabot}{https://github.com/cmu-cabot}} powered by Cartographer~\footnote{\href{https://github.com/cartographer-project/cartographer}{https://github.com/cartographer-project/cartographer}}.

For pedestrian tracking, we first tracked the pedestrians on the overhead camera videos. We found ByteTrack \cite{zhang2021bytetrack} to be very successful in tracking pedestrians in the image space. Upon human verification over our entire data, ByteTrack successfully tracks $95.1\%$ of the pedestrians automatically.

For the automatically tracked labels in pixel space, we needed to convert them into metric space. Each camera video contained a set of tracked trajectories in the image space. We estimated the 3D trajectory coordinates for each pair of 2D trajectories from different cameras and the set of estimated coordinates that resulted in the lowest reprojection error were selected to be the trajectory coordinates in the metric space. 

\subsection{Human Label Verification}

\begin{figure}[]
    \centering
    \includegraphics[scale=0.23]{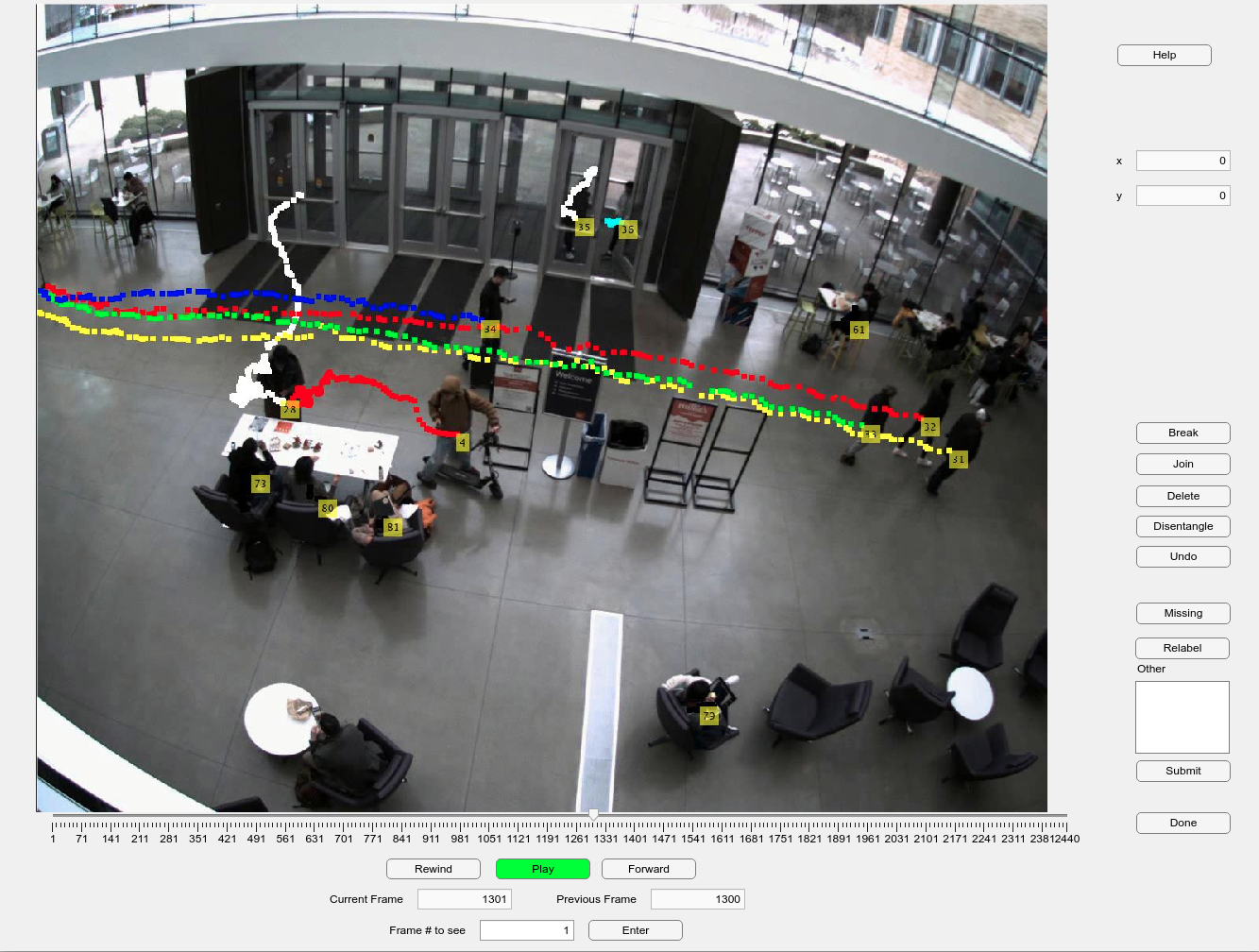}
    \caption{Application interface for the human verification process. It contains a media player and various options to fix tracking errors automatically and manually.}
    \label{fig:hdc_app}
    \vspace{-1em}
\end{figure}

To ensure the label quality of the data, human verification of the tracked trajectories from ByteTack is desired. Semi-autonomous labeling procedures are common in autonomous driving datasets and pedestrian datasets. However, surveying existing pedestrian datasets literature, we noticed that datasets that do contain human-verified metric space labels are often relatively small \cite{ETH, UCY, jrdb, wildtrack}, and large-scale datasets often only use automated tracking pipelines \cite{atc, edinburgh, virat} or do not label surrounding pedestrians \cite{karnan2022scand, diego2022crowdbot}. We attribute this to a lack of tools to streamline the human verification process.

To this end, we designed an open-source web application (Figure~\ref{fig:hdc_app}) using Matlab App Designer. The tool was designed to minimize complete human relabeling of erroneously tracked trajectories. The application contains a media player. When using the application, human labelers only need to watch videos with the automatically tracked trajectories. When an error is noticed, the labeler only needs to indicate to the system the type and location of the error. The system then fixes the errors in the background, and updates the trajectory visualization accordingly. Currently, the application contains the following set of error-fixing options:
\begin{itemize}
    \item \textbf{Break}: Used when ByteTrack incorrectly assigns the same trajectory to two different pedestrians.
    \item \textbf{Join}: Used when two different trajectories actually belong to the same pedestrian.
    \item \textbf{Delete}: Used when a ghost trajectory appears, such as incorrectly tracking an unworn jacket as a pedestrian.
    \item \textbf{Disentangle}: Used when the two trajectories of two pedestrians swapped in the middle, which can happen when one partially occludes the other.
\end{itemize}
The web application also supports undoing previous actions, partial or complete relabeling of trajectories, and labeling missing trajectories. For future work, we are looking at possible platforms to launch the application so that the human verification process can be a crowdsource effort.

Combined with ByteTrack, it took an expert labeler about 30 hours to produce human-verified labels for 375K frames of data, or 10300 trajectories. ByteTrack successfully tracks $95.1\%$ of trajectories. Among the trajectory errors that require human rectification, $0.35\%$ are fixed by ``Break", $1.25\%$ are fixed by ``Join", $0.29\%$ are fixed by ``Delete", $0.89\%$ are fixed by ``Disentangle", $1.21\%$ are fixed by ``Relabel", and $0.48\%$ are fixed by ``Missing".
\section{Dataset Characteristics and Analysis} \label{sec:evaluation}

\subsection{Comparison with Existing Datasets} \label{sec:eval-compare}
\begin{table}[]
\caption{A survey of existing pedestrian datasets on how they incorporate the three components in section \ref{sec:eval-compare}. For component 1, a ``No" means either not human verified or not grounded in metric space. For component 2, TD stands for ``top-down view" and ``E" stands for ``ego-centric view".}
\label{tab:survey}
\begin{center}
\begin{tabular}{c||ccc}
\toprule
Datasets & Comp. 1 & Comp. 2 & Comp. 3\\
&   (metric labels) & (views) & (``robot") \\
\hline
TBD (Ours) & Yes & TD + E & Human + Robot \\
ETH \cite{ETH} & Yes & TD & N/A \\
UCY \cite{UCY} & Yes & TD & N/A \\
Edinburgh Forum \cite{edinburgh} & No & TD & N/A \\
VIRAT \cite{virat} & No & TD & N/A \\
Town Centre \cite{towncentre} & No & TD & N/A \\
Grand Central \cite{grandcentral} & No & TD & N/A \\
CFF \cite{cff} & No & TD & N/A \\
Stanford Drone \cite{stanforddrone} & No & TD & N/A \\
L-CAS \cite{lcas} & No* & E & Robot\\
WildTrack \cite{wildtrack} & Yes & TD & N/A\\
JackRabbot \cite{jrdb} & Yes & E & Robot\\
ATC \cite{atc} & No & TD & N/A\\
TH\"OR \cite{thor} & Yes & TD + E & Robot\\
SCAND \cite{karnan2022scand} & No & E & Robot\\
Crowd-Bot \cite{diego2022crowdbot} & No & E & Human + Robot\\
\bottomrule
\end{tabular}
\end{center}
\vspace{-1.5em}
\end{table}

Compared to existing datasets collected in natural pedestrian-dominant environments, our TBD pedestrian dataset contains three components that greatly enhances the dataset's utility. These components are:

\textbf{Human verified labels grounded in metric space.} ETH \cite{ETH} and UCY \cite{UCY} datasets are the most popular datasets among human behavior analysis papers \cite{rudenko2019-predSurvey}. We believe this is partly because the trajectory labels in these datasets are human verified and are grounded in metric space rather than pixel space (e.g. \cite{stanforddrone} and \cite{towncentre} only contain labels in bounding boxes). Having labels grounded in metric space eliminates the possibility that camera poses might have an effect on the scale of the labels. It also makes the dataset useful for robot navigation related research because robots plan in the metric space rather than pixel space.
    
\textbf{Combination of top-down views and ego-centric views.} Similar to datasets with top-down views, we used top-down views to obtain ground truth trajectory labels for every pedestrian present in the scene. Similar to datasets with ego-centric views, we gathered ego-centric views from a ``robot" to imitate robot perception of human crowds. A dataset that contains both top-down views and ego-centric views will be useful for research projects that rely on ego-centric views. This allows ego-centric inputs to their models, while still having access to ground truth knowledge of the entire scene.

\textbf{Naturalistic human behavior with the presence of a ``robot".} Unlike datasets such as \cite{karnan2022scand}, \cite{diego2022crowdbot}, \cite{jrdb}, and \cite{lcas}, the ``robot" that provides ego-centric view data collection is a cart or a suitcase robot being pushed by human. As mentioned in section \ref{sec:hardware}, doing so reduces the novelty effects from the surrounding pedestrians. Having the ``robot" pushed by humans also ensures safety for the pedestrians and its own motion has less jerk and more humanlike behavior.

As shown in Table \ref{tab:survey}, current datasets only contain at most two of the three components\footnote{*L-CAS dataset does provide human verified labels grounded in the metric space. However, its pedestrian labels do not contain trajectory data.}. A close comparison is the TH\"OR dataset \cite{thor}, but its ego-centric view data are collected by a robot running on predefined trajectories. Additionally, unlike all other datasets in Table \ref{tab:survey}, the TH\"OR dataset is collected in a controlled lab setting rather than in the wild. This injects artificial factors into human behavior. 

\subsection{Dataset Size} \label{sec:eval-stats}

\begin{table}[]
\caption{Dataset comparison statistics, for those with human verified labels grounded in metric space. Numbers in parenthesis are for data that includes the ego-centric view.}
\label{tab:stats}
\begin{center}
\begin{tabular}{c|ccc}
\toprule
Datasets & Duration & \# Trajectories & Label Freq (Hz)\\
\hline
TBD Set 1 & 133 (51) min & 1416 & 60 \\
TBD Set 2 & 626 (213) min & 10300 & 10 \\
ETH \cite{ETH} & 25 min & 650 & 15 \\
UCY \cite{UCY} & 16.5 min & 786 & 2.5 \\
WildTrack \cite{wildtrack} & 200 sec & 313 & 2\\
JackRabbot \cite{jrdb} & 62 min & 260 & 7.5\\
TH\"OR \cite{thor} & 60+ min & 600+ & 100\\
\bottomrule
\end{tabular}
\end{center}

\end{table}

Table \ref{tab:stats} demonstrates the ability of a semi-automatic labeling pipeline to produce large amounts of data. With the aid of an autonomous tracker, humans only need to verify and make occasional corrections on the tracking outcomes instead of locating the pedestrians on every single frame. The data we have collected so far surpassed all other datasets that provide human-verified labels in the metric space in terms of total time and number of pedestrians. We will continue this effort and collect more data for future work.

\subsection{Dataset Statistics}
\begin{table}[]
\caption{Comparison of statistics between our dataset and other datasets according to the methods in \cite{thor}.}
\label{tab:more-stats}
\begin{center}
\begin{tabular}{c|cccc}
\toprule
\multirow{3}{*}{Datasets} & Tracking & Percep. & Motion & Min Dist. \\
 & Duration & Noise & Speed & To Ped.\\
 & [$s$] & [$ms^{-2}$] & [$ms^{-1}$] & [$m$] \\
\hline
TBD Set 2 & $25.6\pm57.1$ & $0.55$ & $0.88\pm0.52$ & $1.25\pm1.44$\\
TH\"OR \cite{thor} & $16.7\pm14.9$ & $0.12$ & $0.81\pm0.49$ & $1.54\pm1.60$\\
ETH \cite{ETH} & $9.4\pm5.4$ & $0.19$ & $1.38\pm0.46$ & $1.33\pm1.39$\\
ATC \cite{atc} & $39.7\pm64.7$ & $0.48$ & $1.04\pm0.46$ & $0.61\pm0.16$\\
Edinburgh  & 
\multirow{2}{*}{$10.1\pm12.7$} & 
\multirow{2}{*}{$0.81$} & 
\multirow{2}{*}{$1.0\pm0.64$} & 
\multirow{2}{*}{$3.97\pm3.5$} \\
\cite{edinburgh} & & & & \\
\bottomrule
\end{tabular}
\end{center}
\vspace{-1em}
\end{table}

Extending the evaluations performed in TH\"OR \cite{thor}, we added the same suite of analysis on Set 2 of our TBD dataset. The evaluation metrics were the following. (1) \textit{Tracking Duration} ($s$): Average time duration of tracked trajectories. (2) \textit{Perception Noise} ($ms^{-2}$): The average absolute acceleration of the trajectories. (3) \textit{Motion Speed} ($ms^{-1}$): Velocities of the trajectories measured in 1 second intervals. (4) \textit{Minimum Distance Between People} ($m$): Minimum Euclidean distance between two closest observed people.

As shown in Table \ref{tab:more-stats}, our dataset has considerable average trajectory duration ($\pm 25.6$) and large variation ($\pm 57.1$), second only to ATC, which has a $900m^2$ coverage. While our dataset has a much smaller coverage, we attribute this to the presence of pedestrians changing navigation goals and static pedestrians in our dataset. Static pedestrians include standing pedestrians having conversations or pedestrians sitting on chairs. Their presence in our dataset often has a long duration, which also causes big variation in this metric. The tracking noise of our system was sub-optimal when compared to ETH and TH\"OR, which is likely due to noisy tracking of the sitting pedestrians. We observed that sitting pedestrians change their body poses frequently, which causes the tracked bounding boxes to change size frequently. We will investigate how to improve this for future work. The motion speeds of our dataset trajectories are lower ($0.88$), which suggests the presence of more static pedestrians. We also have the second-highest variation in motion speed ($\pm0.52$), suggesting that our dataset captures a wide range of pedestrian behavior. From the minimum distance between people, it can be inferred that our dataset captures both dense and sparse population scenarios, as indicated by the middle mean value ($1.25$) among the others and high variance ($\pm 1.44$). Note that \cite{thor} also measures trajectory curvatures, but we noticed that this measurement is heavily affected by how static pedestrians are processed. \cite{thor} does not provide any details on this, so we decided not to evaluate this metric.

\subsection{Behavior Distribution Analysis}

\begin{table}[]
\caption{Trajectory prediction displacement error on ETH/UCY datasets and TBD dataset Set 2. }
\label{tab:pred-stats}
\begin{center}
\begin{tabular}{c|cccc}
\toprule
& \multicolumn{4}{c}{ETH/UCY Dataset} \\
\hline
\multirow{2}{*}{Models} & \multicolumn{2}{c}{Static + Dynamic} & \multicolumn{2}{c}{Dynamic} \\
 & ADE($m$) & FDE($m$) & ADE($m$) & FDE($m$) \\
 \hline
 Social-GAN \cite{Gupta1} &
 0.48 & 0.96 & 0.59 & 1.13 \\
 Trajectron++ \cite{salzmann2020trajectron++} &
 0.27 & 0.49 & 0.35 & 0.65 \\
 AgentFormer \cite{yuan2021agentformer} &
 0.23 & 0.39 & 0.25 & 0.44 \\
 \hline
 & \multicolumn{4}{c}{TBD Set 2} \\
 \hline
  Social-GAN &
 0.36 & 0.72 & 0.64 & 1.30 \\
 Trajectron++ &
 0.16 & 0.28 & 0.43 & 0.83\\
 AgentFormer &
 0.15 & 0.23 & 0.30 & 0.52\\
\bottomrule
\end{tabular}
\end{center}
\vspace{-1em}
\end{table}

\begin{figure*}[]
      \centering
      \includegraphics[scale=0.45]{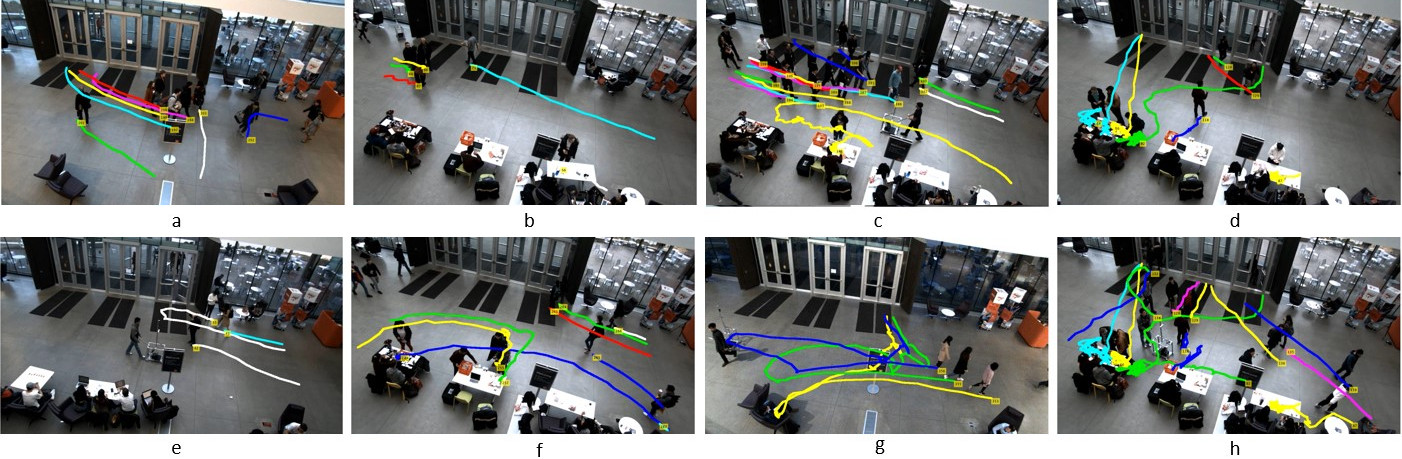}
      \caption{Examples from the TBD Set 1. a) a dynamic group. b) a static conversational group. c) a tour group with 14 pedestrians. d) a pedestrian affecting other pedestrians by asking them to come to the table. e) pedestrians stop and look at their phones. f) two pedestrians change their navigation goals and turn toward the table. g) a group of pedestrians change their navigation goals multiple times. h) a crowded scene where pedestrians are heading in different directions.}
      \label{fig:qual}
      \vspace{-1em}
   \end{figure*}

We additionally leveraged trajectory prediction models to evaluate our dataset. Trajectory prediction models' performance has advanced significantly. We believe these well-trained models can be utilized in the other ways, such as characterizing the variety of pedestrian behavior in datasets. Almost all trajectory prediction models have been tuned and trained on ETH/UCY datasets. Some have additionally made predictions on SDD \cite{stanforddrone} or autonomous driving datasets. Because we were primarily concerned with metric labels and pedestrian environments, we did not evaluate models trained on SDD or autonomous vehicle datasets. We also chose models that largely leverage pedestrian positional data and can work independently without image patch inputs \cite{Sophie} or semantic segmentation \cite{mangalam2021ynet}.

 Our dataset contains simple sessions and challenging sessions. To test if our dataset contains pedestrian behavior outside of ETH/UCY dataset domains, we analyzed the sessions from our dataset with great variety in pedestrian behavior for this evaluation. We selected Social-GAN \cite{Gupta1} as the baseline model and Trajectron++ \cite{salzmann2020trajectron++} and AgentFormer \cite{yuan2021agentformer} as relatively state-of-the-art models. Because the models trained on each of the other four sub-datasets did not perform significantly differently in our dataset, we only report the average \textit{Average Displacement Error} (ADE) and the average \textit{Final Displacement Error} (FDE) across the five models.

We observed that when all pedestrians are concerned, the prediction models all perform better in our dataset (Table~\ref{tab:pred-stats}). We believe it can be attributed to the larger presence of static pedestrians in our datasets compared to ETH/UCY because it is unlikely to yield large errors when predicting future trajectories of static pedestrians. We additionally defined dynamic pedestrians as pedestrians who move at least $1m$ during the prediction window. We included static pedestrians during model inference but only dynamic pedestrians were considered for evaluation. With this, we discovered that all the prediction models' performances degrade. This indicates that the models have encountered more unseen scenarios in our dataset and that the moving pedestrians in our dataset exhibit more diverse navigation behavior and wider behavior distribution compared to the ones in ETH/UCY. 

\subsection{Qualitative Pedestrian Behavior}

Due to the nature of the environment where we collect the data, we observe a mixture of corridor and open-space pedestrian behavior, many of which are rarely seen in other datasets. As shown in Figure \ref{fig:qual}, we observe both static conversation groups and dynamic walking groups. We also observe that some pedestrians change goals mid-navigation. During certain sessions, some pedestrians set up activity areas with tables and chairs to engage with passing pedestrians. This creates more interesting interaction scenarios.
\section{Conclusion}

This paper presents a data collection system that is portable and enables large-scale data collection. This paper also presents a human label verification tool that streamlines the labeling process. Our semi-autonomous pipeline easily produces human-verified labels in order to meet the demands of the large-scale data collected by our hardware. Our systems offer better utility for pedestrian behavior research because our systems consist of human-verified labels grounded in the metric space, a combination of both top-down views and ego-centric views, and a human-pushed cart or robot that approximates naturalistic human motion with a socially aware robot. Lastly, we present the TBD Pedestrian Dataset we have collected using our system, which not only surpasses the quantity of similar datasets, but also offers unique pedestrian interaction behavior that adds to the qualitative diversity of pedestrian interaction data.

As mentioned earlier, our approach enables additional data collection in a wide range of locations and constraints. Additional data collection and public updates to this initial dataset are planned. We have also discovered additional challenges with our labeling pipeline on static pedestrians, because static pedestrians have long trajectory duration and constantly adjust their body poses, the resulting trajectories can be noisy and escape the labeler's attention when using our tool. For future works, we would also explore expanding the diversity of the labels. Some examples include: adding activity labels indicating whether the pedestrian is walking, talking or sitting; adding static obstacle labels for human-object interaction studies; adding group labels for pedestrian groups; and adding gaze direction and head orientation labels for the onboard high-definition 360 camera.

%\addtolength{\textheight}{-11cm}   % This command serves to balance the column lengths

%%%%%%%%%%%%%%%%%%%%%%%%%%%%%%%%%%%%%%%%%%%%%%%%%%%%%%%%%%%%%%%%%%%%%%%%%%%%%%%%

%%%%%%%%%%%%%%%%%%%%%%%%%%%%%%%%%%%%%%%%%%%%%%%%%%%%%%%%%%%%%%%%%%%%%%%%%%%%%%%%

%%%%%%%%%%%%%%%%%%%%%%%%%%%%%%%%%%%%%%%%%%%%%%%%%%%%%%%%%%%%%%%%%%%%%%%%%%%%%%%%

\section*{ACKNOWLEDGMENTS}
This work was supported by the National Science Foundation (IIS-1734361), National Institute on Disability, Independent Living, and Rehabilitation Research (NIDILRR 90DPGE0003), Office of Naval Research (ONR N00014-18-1-2503), and Shimizu Corporation. We would like to thank colleagues in the Tepper School of Business for their assistance on data collection logistics.

%%%%%%%%%%%%%%%%%%%%%%%%%%%%%%%%%%%%%%%%%%%%%%%%%%%%%%%%%%%%%%%%%%%%%%%%%%%%%%%%
{
\bibliographystyle{IEEEtran}
\bibliography{IEEEabrv}

\begin{thebibliography}{10}
\providecommand{\url}[1]{#1}
\csname url@rmstyle\endcsname
\providecommand{\newblock}{\relax}
\providecommand{\bibinfo}[2]{#2}
\providecommand\BIBentrySTDinterwordspacing{\spaceskip=0pt\relax}
\providecommand\BIBentryALTinterwordstretchfactor{4}
\providecommand\BIBentryALTinterwordspacing{\spaceskip=\fontdimen2\font plus
\BIBentryALTinterwordstretchfactor\fontdimen3\font minus \fontdimen4\font\relax}
\providecommand\BIBforeignlanguage[2]{{%
\expandafter\ifx\csname l@#1\endcsname\relax
\typeout{** WARNING: IEEEtran.bst: No hyphenation pattern has been}%
\typeout{** loaded for the language `#1'. Using the pattern for}%
\typeout{** the default language instead.}%
\else
\language=\csname l@#1\endcsname
\fi
#2}}

\bibitem{Alahi1}
A.~Alahi, K.~Goel, V.~Ramanathan, A.~Robicquet, L.~Fei-Fei, and S.~Savarese, ``Social lstm: Human trajectory prediction in crowded spaces,'' in \emph{Proc. IEEE Comput. Soc. Conf. Comput. Vis. Pattern Recognit.}, June 2016, pp. 961--971.

\bibitem{Gupta1}
A.~Gupta, J.~Johnson, L.~Fei-Fei, S.~Savarese, and A.~Alahi, ``Social gan: Socially acceptable trajectories with generative adversarial networks,'' in \emph{Proc. IEEE Comput. Soc. Conf. Comput. Vis. Pattern Recognit.}, June 2018, pp. 2255--2264.

\bibitem{ivanovic-trajectron}
B.~Ivanovic and M.~Pavone, ``The trajectron: Probabilistic multi-agent trajectory modeling with dynamic spatiotemporal graphs,'' in \emph{Proc. IEEE/CVF International Conf. on Comput. Vis.}, 2019, pp. 2375--2384.

\bibitem{kitani-2012}
K.~M. Kitani, B.~D. Ziebart, J.~A. Bagnell, and M.~Hebert, ``Activity forecasting,'' in \emph{Comput. Vis. -- ECCV 2012}, 2012, pp. 201--214.

\bibitem{liang2020garden}
J.~Liang, L.~Jiang, K.~Murphy, T.~Yu, and A.~Hauptmann, ``The garden of forking paths: Towards multi-future trajectory prediction,'' in \emph{Proc. IEEE Conf. on Comput. Vis. and Pattern Recognit.}, 2020.

\bibitem{okal-IRL}
B.~Okal and K.~O. Arras, ``Learning socially normative robot navigation behaviors with bayesian inverse reinforcement learning,'' in \emph{2016 IEEE International Conf. on Robotics and Automation (ICRA)}.\hskip 1em plus 0.5em minus 0.4em\relax IEEE, 2016, pp. 2889--2895.

\bibitem{kretzschmar_ijrr16}
H.~Kretzschmar, M.~Spies, C.~Sprunk, and W.~Burgard, ``Socially compliant mobile robot navigation via inverse reinforcement learning,'' \emph{The International Journal of Robotics Research}, vol.~35, no.~11, pp. 1289--1307, 2016.

\bibitem{biswas2021socnavbench}
A.~Biswas, A.~Wang, G.~Silvera, A.~Steinfeld, and H.~Admoni, ``Socnavbench: A grounded simulation testing framework for evaluating social navigation,'' \emph{arXiv preprint arXiv:2103.00047}, 2021.

\bibitem{karnan2022scand}
H.~Karnan, A.~Nair, X.~Xiao, G.~Warnell, S.~Pirk, A.~Toshev, J.~Hart, J.~Biswas, and P.~Stone, ``Socially compliant navigation dataset (scand): A large-scale dataset of demonstrations for social navigation,'' \emph{IEEE Robotics and Automation Letters}, vol.~7, no.~4, pp. 11\,807--11\,814, 2022.

\bibitem{diego2022crowdbot}
D.~Paez-Granados, Y.~He, D.~Gonon, D.~Jia, B.~Leibe, K.~Suzuki, and A.~Billard, ``Pedestrian-robot interactions on autonomous crowd navigation: Reactive control methods and evaluation metrics,'' in \emph{2022 IEEE/RSJ International Conference on Intelligent Robots and Systems (IROS)}, 2022, pp. 149--156.

\bibitem{atc}
D.~Br{\v{s}}{\v{c}}i{\'c}, T.~Kanda, T.~Ikeda, and T.~Miyashita, ``Person tracking in large public spaces using 3-d range sensors,'' \emph{IEEE Trans. on Human-Machine Syst.}, vol.~43, no.~6, pp. 522--534, 2013.

\bibitem{edinburgh}
B.~Majecka, ``Statistical models of pedestrian behaviour in the forum,'' \emph{Master's thesis, School of Informatics, University of Edinburgh}, 2009.

\bibitem{stanforddrone}
A.~Robicquet, A.~Sadeghian, A.~Alahi, and S.~Savarese, ``Learning social etiquette: Human trajectory understanding in crowded scenes,'' in \emph{Comput. Vis. -- ECCV 2016}, B.~Leibe, J.~Matas, N.~Sebe, and M.~Welling, Eds., 2016, pp. 549--565.

\bibitem{virat}
S.~Oh, A.~Hoogs, A.~Perera, N.~Cuntoor, C.-C. Chen, J.~T. Lee, S.~Mukherjee, J.~Aggarwal, H.~Lee, L.~Davis, \emph{et~al.}, ``A large-scale benchmark dataset for event recognit. in surveillance video,'' in \emph{CVPR 2011}.\hskip 1em plus 0.5em minus 0.4em\relax IEEE, 2011, pp. 3153--3160.

\bibitem{ETH}
S.~Pellegrini, A.~Ess, K.~Schindler, and L.~van Gool, ``You'll never walk alone: Modeling social behavior for multi-target tracking,'' in \emph{Proc. IEEE Int. Conf. Comput. Vis.}, Sept 2009, pp. 261--268.

\bibitem{UCY}
A.~Lerner, Y.~Chrysanthou, and D.~Lischinski, ``Crowds by example,'' \emph{Comput. Graph. Forum}, vol.~26, no.~3, pp. 655--664, 2007.

\bibitem{jrdb}
R.~Martin-Martin, M.~Patel, H.~Rezatofighi, A.~Shenoi, J.~Gwak, E.~Frankel, A.~Sadeghian, and S.~Savarese, ``Jrdb: A dataset and benchmark of egocentric robot visual perception of humans in built environments,'' \emph{IEEE Trans. Pattern Anal. Mach. Intell.}, 2021.

\bibitem{Trautman1}
P.~Trautman and A.~Krause, ``Unfreezing the robot: Navigation in dense, interacting crowds,'' in \emph{Proc. IEEE/RSJ Int. Conf. Intell. Robot. Syst.}, Oct 2010, pp. 797--803.

\bibitem{sun2021move}
M.~Sun, F.~Baldini, P.~Trautman, and T.~Murphey, ``Move beyond trajectories: Distribution space coupling for crowd navigation,'' \emph{arXiv preprint arXiv:2106.13667}, 2021.

\bibitem{nishimura2020risk}
H.~Nishimura, B.~Ivanovic, A.~Gaidon, M.~Pavone, and M.~Schwager, ``Risk-sensitive sequential action control with multi-modal human trajectory forecasting for safe crowd-robot interaction,'' in \emph{2020 IEEE/RSJ International Conf. on Intell. Robots and Syst. (IROS)}.\hskip 1em plus 0.5em minus 0.4em\relax IEEE, 2020, pp. 11\,205--11\,212.

\bibitem{mavrogiannis_etal2021-core-challenges}
C.~{Mavrogiannis}, F.~{Baldini}, A.~{Wang}, D.~{Zhao}, P.~{Trautman}, A.~{Steinfeld}, and J.~{Oh}, ``{Core Challenges of Social Robot Navigation: A Survey},'' \emph{arXiv e-prints}, p. arXiv:2103.05668, Mar. 2021.

\bibitem{nuscenes}
H.~Caesar, V.~Bankiti, A.~H. Lang, S.~Vora, V.~E. Liong, Q.~Xu, A.~Krishnan, Y.~Pan, G.~Baldan, and O.~Beijbom, ``nuscenes: A multimodal dataset for autonomous driving,'' in \emph{CVPR}, 2020.

\bibitem{Cityscapes}
M.~Cordts, M.~Omran, S.~Ramos, T.~Rehfeld, M.~Enzweiler, R.~Benenson, U.~Franke, S.~Roth, and B.~Schiele, ``The cityscapes dataset for semantic urban scene understanding,'' in \emph{Proc. of the IEEE Conference on Computer Vision and Pattern Recognition (CVPR)}, 2016.

\bibitem{zhang2021bytetrack}
Y.~Zhang, P.~Sun, Y.~Jiang, D.~Yu, Z.~Yuan, P.~Luo, W.~Liu, and X.~Wang, ``Bytetrack: Multi-object tracking by associating every detection box,'' \emph{arXiv preprint arXiv:2110.06864}, 2021.

\bibitem{Sophie}
A.~Sadeghian, V.~Kosaraju, A.~Sadeghian, N.~Hirose, H.~Rezatofighi, and S.~Savarese, ``Sophie: An attentive gan for predicting paths compliant to social and physical constraints,'' in \emph{Proc. IEEE/CVF Conf. on Comput. Vis. and Pattern Recognit. (CVPR)}, June 2019.

\bibitem{Social-STGCNN}
A.~Mohamed, K.~Qian, M.~Elhoseiny, and C.~Claudel, ``Social-stgcnn: A social spatio-temporal graph convolutional neural network for human trajectory prediction,'' in \emph{Proc. IEEE/CVF Conf. on Comput. Vis. and Pattern Recognit. (CVPR)}, June 2020.

\bibitem{wang-split-merge}
A.~Wang and A.~Steinfeld, ``Group split and merge prediction with 3{D} convolutional networks,'' \emph{IEEE Trans. Robot. Autom.}, vol.~5, no.~2, pp. 1923--1930, 2020.

\bibitem{rudenko2019-predSurvey}
A.~Rudenko, L.~Palmieri, M.~Herman, K.~M. Kitani, D.~M. Gavrila, and K.~O. Arras, ``Human motion trajectory prediction: A survey,'' \emph{arXiv preprint arXiv:1905.06113}, 2019.

\bibitem{Everett18_IROS}
M.~Everett, Y.~F. Chen, and J.~P. How, ``Motion planning among dynamic, decision-making agents with deep reinforcement learning,'' in \emph{IEEE/RSJ International Conf. on Intell. Robots and Syst. (IROS)}, Madrid, Spain, Sept. 2018.

\bibitem{chen2019crowd}
C.~Chen, Y.~Liu, S.~Kreiss, and A.~Alahi, ``Crowd-robot interaction: Crowd-aware robot navigation with attention-based deep reinforcement learning,'' in \emph{Proc. IEEE International Conf. on Robotics and Automation (ICRA)}, 2019, pp. 6015--6022.

\bibitem{Chen-gaze-learn}
Y.~Chen, C.~Liu, B.~E. Shi, and M.~Liu, ``Robot navigation in crowds by graph convolutional networks with attention learned from human gaze,'' \emph{IEEE Trans. Robot. Autom.}, vol.~5, no.~2, pp. 2754--2761, 2020.

\bibitem{Tai-IRL}
L.~Tai, J.~Zhang, M.~Liu, and W.~Burgard, ``Socially compliant navigation through raw depth inputs with generative adversarial imitation learning,'' in \emph{2018 IEEE International Conf. on Robotics and Automation (ICRA)}, 2018, pp. 1111--1117.

\bibitem{gao2021evaluation}
Y.~Gao and C.-M. Huang, ``Evaluation of socially-aware robot navigation,'' \emph{Frontiers in Robotics and AI}, p. 420, 2021.

\bibitem{trautmanijrr}
P.~Trautman, J.~Ma, R.~M. Murray, and A.~Krause, ``Robot navigation in dense human crowds: Statistical models and experimental studies of human-robot cooperation,'' \emph{International Journal of Robotics Research}, vol.~34, no.~3, pp. 335--356, 2015.

\bibitem{cao2019dynamic}
C.~Cao, P.~Trautman, and S.~Iba, ``Dynamic channel: A planning framework for crowd navigation,'' in \emph{2019 International Conf. on Robotics and Automation (ICRA)}.\hskip 1em plus 0.5em minus 0.4em\relax IEEE, 2019, pp. 5551--5557.

\bibitem{wang2022group}
A.~Wang, C.~Mavrogiannis, and A.~Steinfeld, ``Group-based motion prediction for navigation in crowded environments,'' in \emph{Conf. on Robot Learning}.\hskip 1em plus 0.5em minus 0.4em\relax PMLR, 2022, pp. 871--882.

\bibitem{Argoverse}
B.~Wilson, W.~Qi, T.~Agarwal, J.~Lambert, J.~Singh, S.~Khandelwal, B.~Pan, R.~Kumar, A.~Hartnett, J.~Kaesemodel~Pontes, D.~Ramanan, P.~Carr, and J.~Hays, ``Argoverse 2: Next generation datasets for self-driving perception and forecasting,'' in \emph{Proceedings of the Neural Information Processing Systems Track on Datasets and Benchmarks}, J.~Vanschoren and S.~Yeung, Eds., vol.~1.\hskip 1em plus 0.5em minus 0.4em\relax Curran, 2021.

\bibitem{MOTChallenge20}
P.~Dendorfer, H.~Rezatofighi, A.~Milan, J.~Shi, D.~Cremers, I.~Reid, S.~Roth, K.~Schindler, and L.~Leal-Taix\'{e}, ``Mot20: A benchmark for multi object tracking in crowded scenes,'' \emph{arXiv:2003.09003[cs]}, Mar. 2020, arXiv: 2003.09003.

\bibitem{gloor2016pedsim}
C.~Gloor, ``Pedsim: Pedestrian crowd simulation,'' \emph{URL http://pedsim. silmaril. org}, vol.~5, no.~1, 2016.

\bibitem{tsoi2020sean}
N.~Tsoi, M.~Hussein, J.~Espinoza, X.~Ruiz, and M.~V{\'a}zquez, ``Sean: Social environment for autonomous navigation,'' in \emph{Proc. 8th International Conf. on Human-Agent Interaction}, 2020, pp. 281--283.

\bibitem{mavrogiannis2021core}
C.~Mavrogiannis, F.~Baldini, A.~Wang, D.~Zhao, P.~Trautman, A.~Steinfeld, and J.~Oh, ``Core challenges of social robot navigation: A survey,'' \emph{arXiv preprint arXiv:2103.05668}, 2021.

\bibitem{cabot2023kuribayashi}
M.~Kuribayashi, T.~Ishihara, D.~Sato, J.~Vongkulbhisal, K.~Ram, S.~Kayukawa, H.~Takagi, S.~Morishima, and C.~Asakawa, ``Pathfinder: Designing a map-less navigation system for blind people in unfamiliar buildings,'' in \emph{Proceedings of the 2023 CHI Conference on Human Factors in Computing Systems}, ser. CHI '23.\hskip 1em plus 0.5em minus 0.4em\relax New York, NY, USA: Association for Computing Machinery, 2023.

\bibitem{brvsvcic2015escaping}
D.~Br{\v{s}}{\v{c}}i{\'c}, H.~Kidokoro, Y.~Suehiro, and T.~Kanda, ``Escaping from children's abuse of social robots,'' in \emph{Proc. of the tenth annual acm/ieee international Conf. on human-robot interaction}, 2015, pp. 59--66.

\bibitem{lcas}
Z.~Yan, T.~Duckett, and N.~Bellotto, ``Online learning for human classification in 3d lidar-based tracking,'' in \emph{2017 IEEE/RSJ International Conf. on Intell. Robots and Syst. (IROS)}.\hskip 1em plus 0.5em minus 0.4em\relax IEEE, 2017, pp. 864--871.

\bibitem{thor}
A.~Rudenko, T.~P. Kucner, C.~S. Swaminathan, R.~T. Chadalavada, K.~O. Arras, and A.~J. Lilienthal, ``Th{\"o}r: Human-robot navigation data collection and accurate motion trajectories dataset,'' \emph{IEEE Trans. Robot. Autom.}, vol.~5, no.~2, pp. 676--682, 2020.

\bibitem{wildtrack}
T.~Chavdarova, P.~Baqu{\'e}, S.~Bouquet, A.~Maksai, C.~Jose, T.~Bagautdinov, L.~Lettry, P.~Fua, L.~Van~Gool, and F.~Fleuret, ``Wildtrack: A multi-camera hd dataset for dense unscripted pedestrian detection,'' in \emph{Proc. IEEE Conf. on Comput. Vis. and Pattern Recognit.}, 2018, pp. 5030--5039.

\bibitem{towncentre}
B.~Benfold and I.~Reid, ``Stable multi-target tracking in real-time surveillance video,'' in \emph{CVPR 2011}.\hskip 1em plus 0.5em minus 0.4em\relax IEEE, 2011, pp. 3457--3464.

\bibitem{grandcentral}
B.~Zhou, X.~Wang, and X.~Tang, ``Understanding collective crowd behaviors: Learning a mixture model of dynamic pedestrian-agents,'' in \emph{2012 IEEE Conf. on Comput. Vis. and Pattern Recognit.}\hskip 1em plus 0.5em minus 0.4em\relax IEEE, 2012, pp. 2871--2878.

\bibitem{cff}
A.~Alahi, V.~Ramanathan, and L.~Fei-Fei, ``Socially-aware large-scale crowd forecasting,'' in \emph{Proc. IEEE Conf. on Comput. Vis. and Pattern Recognit.}, 2014, pp. 2203--2210.

\bibitem{salzmann2020trajectron++}
T.~Salzmann, B.~Ivanovic, P.~Chakravarty, and M.~Pavone, ``Trajectron++: Dynamically-feasible trajectory forecasting with heterogeneous data,'' in \emph{Computer Vision--ECCV 2020: 16th European Conference, Glasgow, UK, August 23--28, 2020, Proceedings, Part XVIII 16}.\hskip 1em plus 0.5em minus 0.4em\relax Springer, 2020, pp. 683--700.

\bibitem{yuan2021agentformer}
Y.~Yuan, X.~Weng, Y.~Ou, and K.~M. Kitani, ``Agentformer: Agent-aware transformers for socio-temporal multi-agent forecasting,'' in \emph{Proceedings of the IEEE/CVF International Conference on Computer Vision}, 2021, pp. 9813--9823.

\bibitem{mangalam2021ynet}
K.~Mangalam, Y.~An, H.~Girase, and J.~Malik, ``From goals, waypoints \& paths to long term human trajectory forecasting,'' in \emph{Proceedings of the IEEE/CVF International Conference on Computer Vision}, 2021, pp. 15\,233--15\,242.

\end{thebibliography}
}

\end{document}